# Does System One Transfer? Calibrated Typed-Decision Models, Label Budgets, and Translation for Indonesian Regional Languages

**Romi Fadli*, Rusdah, Wahyu Ramadani Putri Utami, Zulfa Malika Amana**
*Faculty of Information Technology, Universitas Budi Luhur, Jakarta, Indonesia*
**Corresponding author: Romi Fadli (2511600872@student.budiluhur.ac.id)*

*Rusdah: rusdah@budiluhur.ac.id; Wahyu Ramadani Putri Utami: 2611600301@student.budiluhur.ac.id; Zulfa Malika Amana: 2611600293@student.budiluhur.ac.id*

**Abstract** Calibrated typed-decision ("System One") models return a label and a probability in a single forward pass at a small fraction of the cost of large language models (LLMs), yet their evaluation has been almost entirely English. We report a pre-registered evaluation of the commercial Jev model and the open Laya-multilingual model on Indonesian regional languages, using the parallel NusaX sentiment (12 languages) and NusaTranslation sentiment and emotion (7 languages) test sets. Jev is more often correct on sentences that are lexically closer to Indonesian in all three datasets, and presenting the human Indonesian translation instead raises its accuracy by up to 37.2 points, most in languages distant from Indonesian. Its expected calibration error does not grow on regional languages, but its ability to rank its own errors degrades in all three datasets. Retrained classifiers need more than 500 labels per language to match zero-shot Jev in 9 of 10 NusaX regional languages. A confidence-gated cascade that sends Jev's 10–30% least confident decisions to a local IndoBERT classifier matches or exceeds the better of the two systems, and routing to an LLM instead matches that LLM at 24–77% lower API cost. Of six further commercial LLMs (three Claude and three GPT models), five are 4–14 points more accurate than Jev, at several hundred times the API cost where it is reported, while the smallest (Claude Haiku) is less accurate. We release the pre-registration, code and all predictions.

## 1 Introduction

Many language-processing tasks in software are not generation problems but decisions: a review is positive or negative, a message expresses anger or fear, a ticket goes to one queue or another. Large language models (LLMs) solve such decisions by generating text that is then parsed, which costs seconds and cents per item. A new class of *typed-decision* models, marketed as "System One" models after the fast, intuitive mode of human cognition [1], instead returns a label from a fixed set of options together with a probability, in one non-autoregressive forward pass. The commercial Jev model of TypeSafe AI [2], [3] and the open-weight Laya family [4] are the first widely available instances. Because their answers are always well-formed and come with a probability, they are attractive as the cheap first layer of a text-classification pipeline, provided that the probability can be trusted.

Independent evaluations of these models appeared within days of their release, but almost all of them are in English. Ibrahim and Zaki [5] compared Jev with 19 LLMs on 18 computational-social-science tasks and found it behind the best LLM but better calibrated than most. Deußer et al. [6] benchmarked Jev on 37 datasets, including the 122-language Belebele reading-comprehension set, and reported that it degrades on low-resource languages. The vendor documentation itself states that the model is primarily tuned on English and should be validated for other languages [3]. What such a model offers the speakers of a low-resource language, and when a team would do better to label data and train a small model of its own, remains unmeasured.

Indonesia makes these questions concrete. It has more than 700 living languages [7], most of them with little digital text [8], and a growing set of human-curated benchmarks [9], [10]. Two of these benchmarks are *parallel*: every test sentence exists in each regional language and in Indonesian, written or translated by native speakers. Parallel data make it possible to compare the same content across languages, and to ask what a model gains when the same sentence is presented in the national language instead.

We use this design to evaluate Jev and Laya-multilingual on the NusaX sentiment benchmark (12 languages) and the NusaTranslation sentiment and emotion benchmarks (7 languages each), against retrained classical and transformer classifiers, a logprob-calibrated LLM and a ladder of frontier LLMs. All hypotheses, tests and the stopping rule were registered before the test sets were opened. We address four research questions:

**RQ1.** How does the zero-shot accuracy of a typed-decision model depend on the closeness of a regional language to Indonesian, and how much is lost by not presenting the Indonesian translation instead?

**RQ2.** Do the model's probabilities remain calibrated, and useful for abstaining on uncertain items, on regional languages?

**RQ3.** How many labelled examples per language does a retrained classifier need to match the zero-shot model?

**RQ4.** Can routing the model's least confident decisions to a local classifier or to an LLM improve the cost–accuracy trade-off, and how does the model compare with frontier LLMs?

Our contributions are the following:

1. **A pre-registered, parallel evaluation.** To our knowledge, this is the first evaluation of typed-decision models on Indonesian regional languages. It covers 12 NusaX and 7 NusaTranslation language versions, paired statistical tests on identical items, and every model arm that the research questions refer to.
2. **A sentence-level resource gradient and the cost of not translating.** Beyond the language-level degradation reported by Deußer et al. [6], we show that the probability of a correct answer rises with the lexical overlap between a sentence and its Indonesian parallel, and we measure what the human Indonesian translation and machine translation add.
3. **Calibration versus discrimination.** Contrary to audits on Russian and Spanish [11] and to our pre-registered prediction, the average calibration of the model does not degrade on regional languages; its ability to rank its own errors does.
4. **Label budgets and local cascades.** Learning curves of retrained classifiers on identical, nested training subsets show how many labels are needed to match the zero-shot model, and a cascade to a *local* classifier, with its threshold chosen on validation data, shows how the two can be combined. A ladder of seven commercial LLMs (DeepSeek, three Claude and three GPT models) places these results on a cost–accuracy frontier.

Sect. 2 reviews related work. Sect. 3 describes the data, systems, metrics and pre-registered analysis. Sect. 4 reports and discusses the results, and Sect. 5 concludes.

## 2 Related Work

### 2.1 Typed-decision models and their evaluation

Jev answers three question types, a choice among options, a score on a rubric and a yes/no "noul" probability, for a text state, and reports calibrated confidences [2]. Laya-multilingual is an open 322M-parameter decision model built on the mmBERT encoder [4], [12]; its model card reports near-chance zero-shot accuracy on many languages and recommends fine-tuning. Early evaluations covered computational social science [5] and a general 37-dataset benchmark [6]. Sun et al. [13] showed that the decision head follows the option name more than the rubric bound to it, which motivates keeping option names identical across all our arms. Outside English, a paired Russian–English audit found lower accuracy and a threefold calibration error on Russian [11]. None of these studies covers an Indonesian language, retrains baselines with varying label budgets, or analyses calibration per language on parallel data.

### 2.2 NLP for Indonesian regional languages

Indonesian NLP has benchmark suites for the national language [14], [15], and resources for the regional languages are growing [7]. NusaX [9] provides sentiment data in ten regional languages, Indonesian and English, and NusaWrites [10] provides the NusaTranslation sentiment and emotion data. The baselines of both papers vary widely across languages, in line with broader multilingual evaluations of generative models [16]. Transfer to such languages depends on pretraining data and on similarity to better-resourced languages [17].

### 2.3 Zero-shot models versus trained classifiers

For sentiment, LLMs approach fine-tuned models on simple polarity but lag on finer tasks [18], and small fine-tuned models still outperform zero-shot generative models on many classification tasks [19]. How many labelled examples a prompt is worth has been measured for English [20]. Commercial sentiment services have been compared in this journal [21], but not for regional languages or against retrained models with controlled label budgets.

### 2.4 Calibration and selective prediction across languages

Calibration is commonly measured by the expected calibration error (ECE) [22], [23] and by proper scoring rules such as the Brier score [24]. Massively multilingual models are worse calibrated on low-resource languages [25], the basis of our prediction for Jev. LLM confidence can be read from token probabilities [26] or stated verbally [27]. Selective prediction evaluates whether confidence ranks correct above incorrect answers, summarised by the area under the risk–coverage curve (AURC) [28]; a model can be well calibrated on average and still rank its errors poorly.

### 2.5 Translate-test and model cascades

Translating the input into a high-resource language before classifying it (translate-test) remains a strong baseline [29], and NLLB covers several Indonesian regional languages [30]. Cascades send only uncertain inputs to a more expensive model [31]. For typed-decision models, cascades from Jev to LLMs have been proposed for text annotation [5] and for LLM-as-a-judge evaluation [32].

### 2.6 How this work differs

Our study differs from this work in four ways. First, it evaluates typed-decision models on Indonesian regional languages, with parallel test sets that allow sentence-level analysis. Second, it retrains classical, transformer and decision-model baselines on identical nested subsets with three seeds, so that label budgets are measured rather than quoted. Third, it separates calibration from discrimination per language. Fourth, its cascade routes to a local classifier, with the operating point fixed on validation data before the test results are seen. All confirmatory analyses were pre-registered.

## 3 Materials and Methods

### 3.1 Data

NusaX-senti [9] contains the same 1,000 sentences in ten regional languages (Acehnese, Balinese, Toba Batak, Banjarese, Buginese, Javanese, Madurese, Minangkabau, Ngaju and Sundanese), in Indonesian, from which they were translated by native speakers, and in English, with three sentiment labels. The NusaTranslation data of NusaWrites [10] translate Indonesian sentiment [15] and emotion [14] sentences into regional languages; we use the 7 languages in which every sentence exists (Betawi, Batak, Javanese, Madurese, Makassarese, Minangkabau and Sundanese). We call them NT-senti (two labels) and NT-emot (five labels). Table 1 gives the sizes. "Regional" denotes every language except Indonesian and English. All three datasets are released under CC BY-SA 4.0 and were pinned by checksum.

**Table 1** Datasets. Sizes are sentences per language (train / validation / test) over the parallel languages; the subset is the fixed stratified test sample used for the LLM ladder

| Dataset | Languages | Labels | Per language | Test items | Subset per language (total) |
|---|---|---|---|---|---|
| NusaX-senti | 12 | negative, neutral, positive | 500 / 100 / 400 | 4,800 | 100 (1,200) |
| NT-senti | 7 | negative, positive | 3,400 / 448 / 1,200 | 8,400 | 200 (1,400) |
| NT-emot | 7 | anger, fear, happy, love, sadness | 3,000 / 401 / 1,000 | 7,000 | 200 (1,400) |

*Note.* NT = NusaTranslation. Test items are sentence × language pairs.

### 3.2 Systems compared

Table 2 lists the systems ("arms"). All zero-shot arms receive the same task specification: an English instruction and one short description per label, with the canonical label names as option keys, because option names alone shift the decisions of typed-decision models [13]. Jev was queried through the System One endpoint of OpenRouter; every response came from the snapshot *typesafe/jev-1.13-20260917*. Two question formulations were pre-registered. In A1 (*noul_en*), Jev answers one yes/no question per label and the probabilities are normalised into a distribution; a pilot on the validation split chose this formulation by a rule committed before the first call. Because a normalised set of yes/no probabilities is not Jev's native distribution, calibration (RQ2) uses the native choice question of arm A1-choice (*choice_en*). A4 presents the Indonesian source sentence to Jev, giving an upper bound for "translating perfectly", and A3 first translates the regional sentence into Indonesian with NLLB-200 (distilled, 600M) [30], for the seven NusaX and three NusaTranslation languages that NLLB supports.

D1 is DeepSeek V4.1 Flash [33] without reasoning, called with temperature 1 and token log-probabilities; its prediction is the label with the highest normalised log-probability. The LLM ladder (D3, D4) consists of Claude Haiku 4.5, Sonnet 5 and Opus 5.5 [34] and the OpenAI GPT-6 Luna, GPT-6 Sol and GPT-5.6 Terra models (low reasoning effort), called through their command-line clients (Claude Code and Codex CLI) with a JSON schema that forces a label and a verbalised probability per label. D1 is the primary LLM comparator, because it runs on the full test sets and yields token probabilities whose calibration can be compared with Jev's; the Claude and GPT models form a secondary ladder. To bound cost, the ladder runs on the fixed test subsets of Table 1. Answers that could not be parsed count as wrong; the rate was at most 0.086% for any arm.

**Table 2** Systems (arms) compared

| Arm | System | Input / training | Confidence | Scope |
|---|---|---|---|---|
| A1 | Jev, yes/no per label (*noul_en*) | regional text | model probability | full test |
| A1-choice | Jev, native choice question (*choice_en*) | regional text | model probability | full test |
| A3 | NLLB-200 600M → Jev (*noul_en*) | machine translation to Indonesian | model probability | full test* |
| A4 | Jev (*noul_en*; A4-choice: *choice_en*) | Indonesian source sentence | model probability | full test |
| D1 | DeepSeek V4.1 Flash, no reasoning | regional text | token log-probability | full test |
| D3, D4 | Claude Haiku 4.5 / Sonnet 5 / Opus 5.5; GPT-6 Luna / GPT-6 Sol / GPT-5.6 Terra | regional text | verbalised | subset |
| B1 | Naive Bayes, SVM, logistic regression (BoW / TF-IDF) | trained per language | model probability† | full test |
| B2 | IndoBERT-base fine-tuned | trained per language / pooled | model probability | full test |
| B3 | XLM-R-base fine-tuned | trained per language / pooled | model probability | full test |
| C1 | Laya-multilingual fine-tuned (RLCD) | trained pooled / per language | model probability | full test |
| C2 | Laya-multilingual zero-shot | regional text | model probability | NusaX full; NT subset |

*Note.* * Languages supported by NLLB-200 only (NusaX: ace, ban, bjn, bug, jav, min, sun; NT: jav, min, sun). † No probabilities for the SVM. Calibration is compared only within the same confidence type.

### 3.3 Trained models and learning curves

B1 follows the published NusaX protocol [9], [35]: hyperparameters are selected on the validation split and the model is refitted on training plus validation data; logistic regression on bag-of-words features (B1-LR) is the primary trained comparator. B2 (IndoBERT-base [14]) and B3 (XLM-R-base [36]) are fine-tuned with Adam at a constant learning rate for at most 100 epochs, with early stopping after three epochs without improvement in validation macro-F1 and a maximum length of 256 tokens. Their learning rate and batch size were chosen once, by a 20-point grid on the Indonesian NusaX validation split (learning rate and batch size, B2: $5 \times 10^{-6}$ and 16; B3: $5 \times 10^{-6}$ and 8). C1 fine-tunes Laya-multilingual with the official RLCD recipe of its authors [4], adapted to one GPU: a policy-gradient loss against a strictly proper scoring rule [24] plus cross-entropy, four epochs, an effective batch of 64 through gradient accumulation, one-hot gold targets, and a temperature fitted on the validation split. No Jev output was ever used as a training label, as the provider's terms prohibit it [37].

Each trained arm is run per language (trained on that language only) and, where pre-registered, pooled over all languages of a dataset, with three seeds. For RQ3, the per-language models are also trained on nested, stratified subsets of $n \in \{10, 25, 50, 100, 250, 500\}$ labelled sentences (up to 1,000 for NusaTranslation). The subsets use the same sentence identifiers in every language and for every model, so the learning curves are paired; B1 re-selects its hyperparameters for every subset. In total, 1,270 fine-tuning runs were completed on Tesla T4 GPUs (24.3 GPU-hours) with no failed run; the zero-shot arms produced 102,400 test predictions. The API cost of all Jev arms was USD 0.98 and that of D1 USD 2.04 at list prices.

### 3.4 Metrics

Macro-F1 is the primary measure of accuracy; accuracy is reported for paired tests. Calibration is measured by the ECE of the top-class confidence [22], [23] with 15 equal-width bins (equal-mass bins as a check), and by the multiclass Brier score, a strictly proper scoring rule [24]. For $N$ items grouped into 15 bins $B_1, \ldots, B_{15}$ by confidence, where acc and conf denote the accuracy and the mean confidence of a bin,

$$\mathrm{ECE} = \Sigma_{b=1..15} \left(|B_b| / N\right) \cdot |\mathrm{acc}(B_b) - \mathrm{conf}(B_b)| \quad (1)$$

Eq. (1) weights each bin's gap between accuracy and confidence by its share of the items. Selective prediction is measured by the AURC [28]. With the items sorted by decreasing confidence, and risk(k) the error rate among the k most confident items,

$$\mathrm{AURC} = (1 / N) \cdot \Sigma_{k=1..N} \mathrm{risk}(k) \quad (2)$$

A lower AURC (Eq. (2)) means that the confidence ranks correct answers above errors. Cost is the API cost per 1,000 decisions, and latency the time per decision; the trained arms have no API cost, and their latency was measured separately on a laptop GPU (see Sect. 4.5).

### 3.5 Pre-registration and statistical analysis

Hypotheses, tests, correction families and the analysis code were committed and tagged (*prereg-v1*) before the test splits were opened; the validation pilot was used only to choose the Jev formulation and to estimate power. The confirmatory hypotheses are:

**H1a.** for A1, the probability of a correct answer increases with the lexical overlap between a regional sentence and its Indonesian parallel (sentence-level chrF);

**H1b.** for each regional language, A4 is more accurate than A1 (confirmed if significant in a majority of the regional languages of every dataset);

**H2a.** the ECE of A1-choice is higher on regional languages than on Indonesian;

**H2b.** the AURC of A1-choice is higher on regional languages than on Indonesian;

**H3a.** the label budget $n*$ at which a trained model matches Jev is smaller where Jev is weaker (NusaX regional languages).

H1a is tested with a logistic mixed model [38], one per dataset, on the regional languages, with crossed random intercepts for sentence *i* and language *j* and the standardised overlap *z*:

$$\text{logit}\ \Pr(\text{correct}_{ij} = 1) = \beta_0 + \beta_1 z_{ij} + u_i + v_j \tag{3}$$

H1a is confirmed if the 95% profile-likelihood interval of $\beta_1$ in Eq. (3) excludes zero; a pre-declared fallback (dropping the language intercept, then a GEE) was not needed. Paired accuracy differences are tested with the exact McNemar test [39], with Holm correction within each pre-registered family (H1b: 24 tests; A1 versus B1-LR and A1 versus D1: 26 tests each). Confidence intervals come from a cluster bootstrap that resamples sentences with all their language versions (B = 2,000) [40]. H2 is confirmed if the interval of the regional-minus-Indonesian difference lies above zero; for NusaTranslation, which has no Indonesian test language, the reference is A4-choice on the Indonesian source sentences. For H3a, $n*$ is the smallest n whose lower 95% bound of macro-F1 reaches Jev's macro-F1, the bound being the 2.5th percentile of 3 seeds × 2,000 bootstrap draws; H3a is tested with a one-sided Spearman correlation between $n*$ and Jev's macro-F1 ("more than 500" ranked highest).

Secondary analyses (the comparisons with B1-LR and D1, translate-test, the LLM ladder and the cascade) are reported with intervals and Holm correction but are not used to claim confirmation. For the cascade, Jev answers every item and its least confident share r is answered again by the fallback model; r was chosen on the validation pilot as the smallest share whose cascade macro-F1 reaches the better of the two single models, and this r was applied once to the test split. Analyses not listed in the pre-registration are labelled exploratory.

### 3.6 Deviations from the pre-registration

All deviations were logged with date and reason before the affected results were computed, except where noted. (i) A data-loading option that would have restricted the "full test" arms to the subsets was fixed before the first test call; the change touched only the query runners. (ii) Quota and network failures of the LLM clients were retried and not counted as model answers; invalid model answers were never re-run. (iii) A configuration bug stopped all B2/B3 grid runs before their first epoch; it was fixed before any of these models was trained. (iv) Before the confirmatory analysis, the tagged script was adapted to the test file names and the pre-declared GEE fallback was implemented; the H3a procedure, which the pre-registration defines in words, was implemented with the pooled-draw bound described above; the same analysis on the validation pilot reproduced the pre-registration dry run exactly. (v) After the results were computed, we found that C1 did not train at $n \leq 50$: with an effective batch of 64, such runs make only four optimiser steps, which the mixed-precision loss scaler most likely skipped, and their predictions were 99–100% identical to the zero-shot C2 model. These points are excluded from the C1 learning curves; this does not change any conclusion, because C1 stays below Jev even at n = 500 (Sect. 4.4). (vi) The operating-point rule of the cascade was fixed in writing before the cascade was computed, as the pre-registration left it open.

## 4 Results and Discussion

### 4.1 Overall performance

Table 3 summarises macro-F1 on the full test sets. On NusaX, zero-shot Jev (0.782) is on par with DeepSeek (0.786) and with the fully fine-tuned encoders (0.783–0.792), and above the classical baseline (0.752). On NusaTranslation, with 3,400 training sentences per language instead of 500, the fine-tuned encoders lead by 4.6 points on sentiment and 1.8 points on emotion. The two Jev formulations differ by less than one point. Laya without fine-tuning (C2) is far below every other arm (0.367–0.590), which confirms its model card: the open decision model needs task data, whereas the commercial one is usable zero-shot.

**Table 3** Macro-F1 on the test sets (trained arms: mean of three seeds), API cost per 1,000 decisions and median latency

| Arm | NusaX-senti | NT-senti | NT-emot | Cost / 1,000 (USD) | Latency p50 (ms) |
|---|---|---|---|---|---|
| A1 Jev (*noul_en*) | 0.782 | 0.781 | 0.629 | 0.015–0.019 | 355–369 |
| A1-choice Jev (*choice_en*) | 0.785 | 0.783 | 0.631 | 0.016–0.019 | 353–371 |
| D1 DeepSeek V4.1 Flash | 0.786 | 0.795 | 0.658 | 0.085–0.119 | 763–853 |
| B1-LR (full training data) | 0.752 | 0.805 | 0.603 | local | — |
| B2 IndoBERT, per language | 0.792 | 0.824 | 0.647 | local | — |
| B2 IndoBERT, pooled | 0.787 | — | — | local | — |
| B3 XLM-R, per language | 0.783 | 0.827 | 0.647 | local | — |
| B3 XLM-R, pooled | 0.806 | — | — | local | — |
| C1 Laya-ft, per language | 0.712 | — | — | local | — |
| C1 Laya-ft, pooled | 0.786 | 0.815 | 0.611 | local | — |
| C2 Laya zero-shot | 0.502 | 0.590‡ | 0.367‡ | local | — |

*Note.* ‡ Fixed test subset (C2 on NusaTranslation). — : regime not run for this dataset. Cost and latency ranges span the three datasets; list prices, calls from Jakarta. Trained arms have no API cost (local latency in Sect. 4.5).

### 4.2 RQ1: closeness to Indonesian and the cost of not translating

H1a is confirmed in all three datasets (Table 4): the log-odds of a correct Jev answer increase by 0.14–0.42 per standard deviation of sentence-level overlap with Indonesian, with every 95% interval above zero, after accounting for sentence difficulty and language. The full mixed model converged in every dataset. Presenting the Indonesian source sentence instead (A4) never lowered accuracy in any language, and the gain is largest where the regional language is least like Indonesian (Fig. 1, left). H1b, which requires a significant gain in a majority of the regional languages of *every* dataset, is nevertheless not confirmed: it holds for NusaX (7 of 10) and NT-emot (7 of 7) but not for NT-senti (3 of 7), where Betawi, Javanese, Minangkabau and Sundanese gain only 1.6–2.2 points, differences that do not survive the Holm correction.

**Table 4** RQ1: sentence-level overlap effect (H1a) and the gain from the Indonesian source sentence (H1b)

| Dataset | $\beta_1$ [95% CI] (H1a) | Items | A4 > A1 (Holm) | A4 − A1 range (points) | Largest gain |
|---|---|---|---|---|---|
| NusaX-senti | 0.33 [0.17, 0.49] | 4,000 | 7 of 10 | 0.5–37.2 | Buginese (+37.2) |
| NT-senti | 0.14 [0.03, 0.24] | 8,400 | 3 of 7 | 1.6–16.2 | Makassarese (+16.2) |
| NT-emot | 0.42 [0.31, 0.55] | 7,000 | 7 of 7 | 5.0–34.1 | Makassarese (+34.1) |

*Note.* $\beta_1$: log-odds per standard deviation of overlap_ind, logistic mixed model with sentence and language intercepts (Eq. (3)). H1b: exact McNemar, Holm over 24 tests.

Compared with the classical baseline trained on all available labels (Fig. 1, right), Jev is significantly more accurate in 9 of 26 language versions and less accurate in 5: Makassarese (NT-emot), Batak (NT-senti), Madurese (NT-senti), Makassarese (NT-senti), Buginese (NusaX-senti). None of these languages is represented in the CC100 pretraining corpus, and they are among those in which the Indonesian source sentence helps most. The pattern is consistent: Jev transfers well to regional languages that share vocabulary with Indonesian or that are represented in multilingual pretraining corpora [36], [8], and poorly to the others.

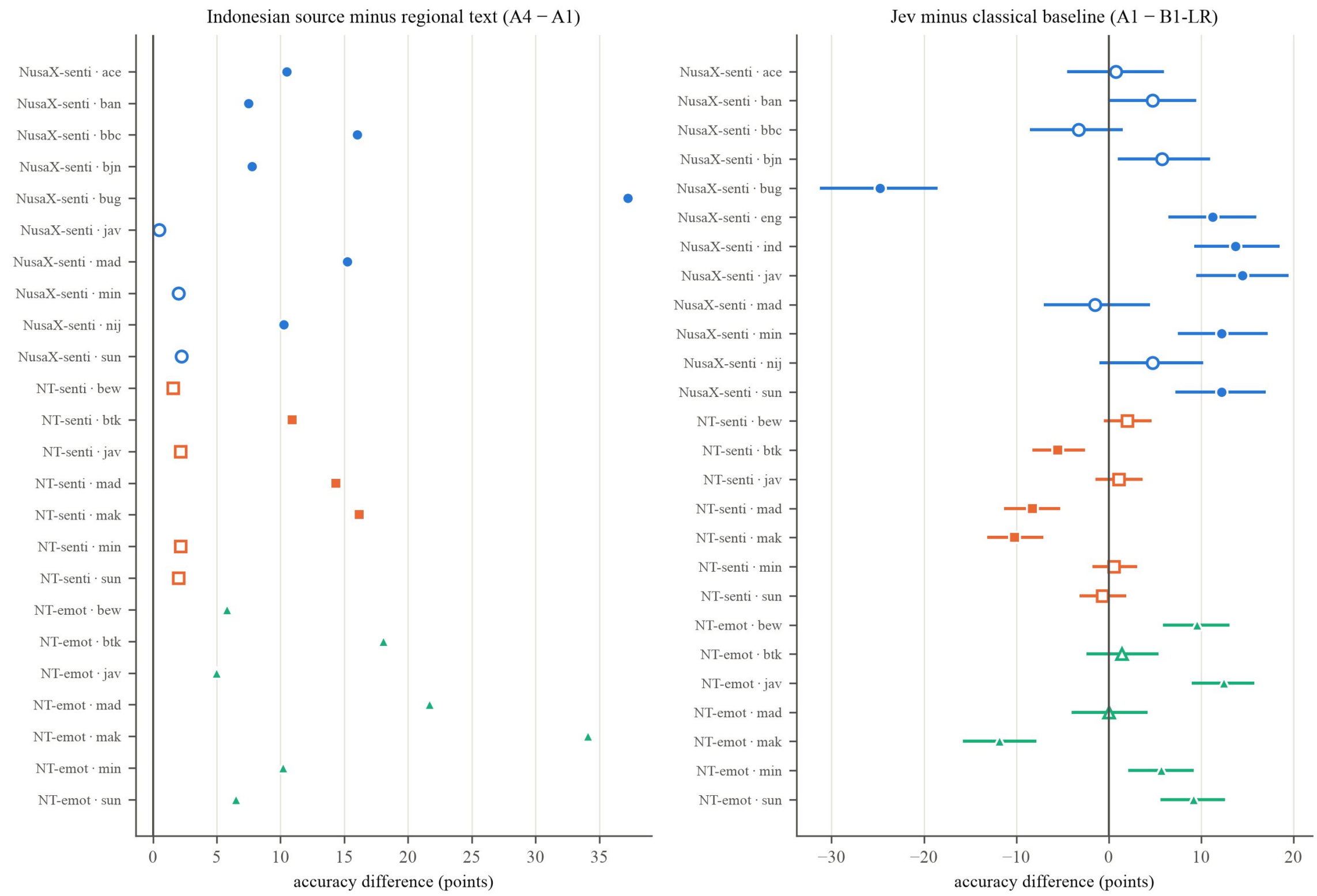

**Fig. 1** Left: accuracy gain when Jev reads the Indonesian source sentence instead of the regional sentence (A4 − A1). Right: Jev minus the classical baseline trained on all labels (A1 − B1-LR) with 95% cluster-bootstrap intervals. Filled markers: Holm-adjusted $p < 0.05$; hollow markers: not significant. Languages are given by their ISO 639-3 codes

### 4.3 RQ2: calibration and selective prediction

H2a is not confirmed. The ECE of Jev's native choice distribution is *lower* on regional languages than on Indonesian in NusaX and NT-senti, and the difference is not significant in NT-emot (Table 5). Mean overconfidence behaves the same way: Jev lowers its confidence on regional text roughly in step with its accuracy. This contradicts our prediction from multilingual encoders [25] and the Russian and Spanish audits [11]. H2b, in contrast, is confirmed in all three datasets: the AURC is higher on regional languages, meaning that on regional text the confidence separates correct from incorrect answers less well. For a practitioner, the probabilities remain honest on average, but a confidence threshold removes fewer errors per abstention than it does for Indonesian.

**Table 5** RQ2: regional minus Indonesian differences for Jev's native choice distribution (A1-choice), with 95% cluster-bootstrap intervals

| Dataset | ΔECE (H2a) | ΔAURC (H2b) | Δ Overconfidence |
|---|---|---|---|
| NusaX-senti | −0.036 [−0.054, −0.018] | 0.058 [0.035, 0.066] | −0.040 [−0.060, −0.020] |
| NT-senti | −0.020 [−0.037, −0.010] | 0.039 [0.030, 0.047] | −0.023 [−0.037, −0.008] |
| NT-emot | 0.010 [−0.013, 0.024] | 0.090 [0.076, 0.105] | 0.017 [−0.002, 0.036] |

*Note.* Positive values: worse on regional languages. For NusaTranslation, the Indonesian reference is Jev on the Indonesian source sentences (A4-choice). Overconfidence = mean confidence − accuracy.

Relative to the logprob-derived distribution of DeepSeek, Jev is better calibrated (Fig. 2). Its ECE is lower by 0.039–0.127 (all intervals below zero), and in an exploratory per-language analysis it is lower in 25 of 25 language versions. DeepSeek is systematically overconfident; its accuracy is similar (Sect. 4.6).

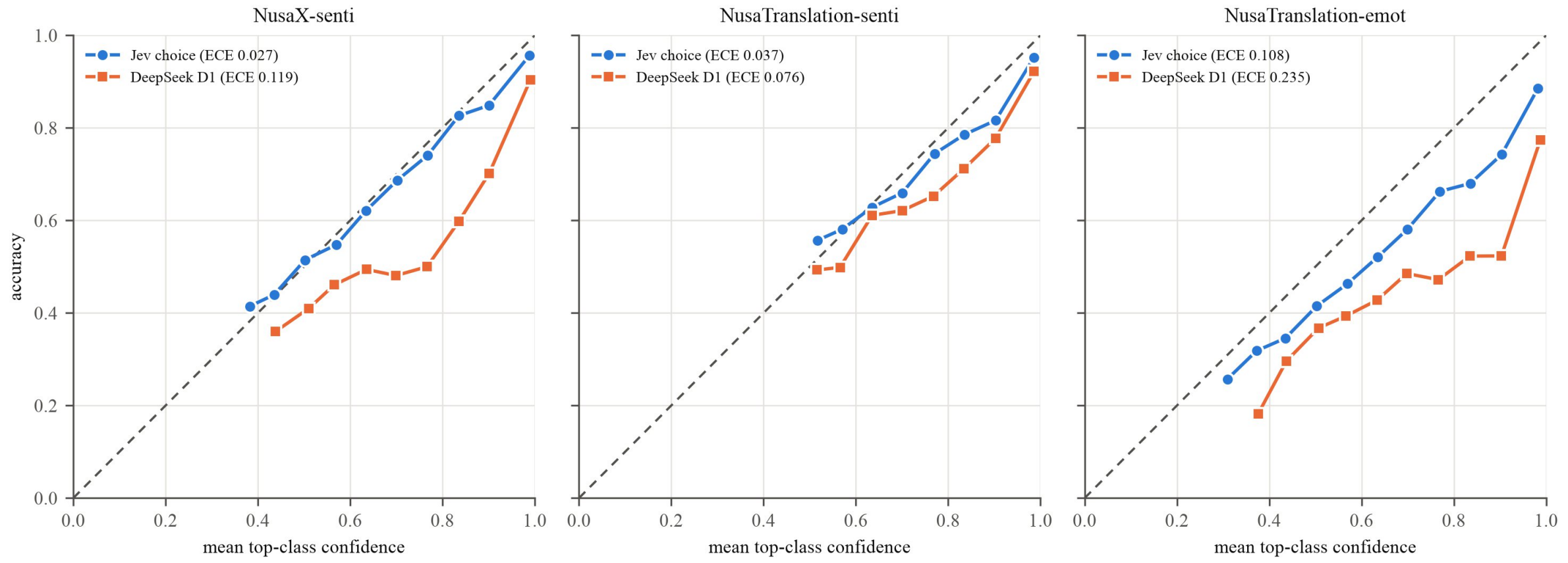

**Fig. 2** Reliability diagrams of Jev's native choice distribution and DeepSeek's logprob distribution on the same items (15 equal-width bins; bins with fewer than ten items omitted)

### 4.4 RQ3: how many labels match zero-shot Jev?

Fig. 3 shows the learning curves and Table 6 the resulting label budgets. With 500 labels per language, the lower bound of the classical baseline still falls short of zero-shot Jev in 9 of 10 NusaX regional languages; IndoBERT reaches it only for Banjarese at n = 500 and Buginese at n = 100, and fine-tuned Laya in no language. The exception is Buginese, where Jev is weak (macro-F1 0.514) and 25 labels suffice. H3a, the prediction that the budget is smaller where Jev is weaker, points in the expected direction but is not confirmed (Spearman $\rho = 0.52$, one-sided $p = 0.061$; IndoBERT $\rho = 0.39$, $p = 0.133$). With nine of ten budgets tied at "more than 500", the test has little power, and the result is better read as the budget itself: for sentiment in most NusaX languages, a team would need several hundred labelled sentences per language before a local model is reliably at least as good as the zero-shot model.

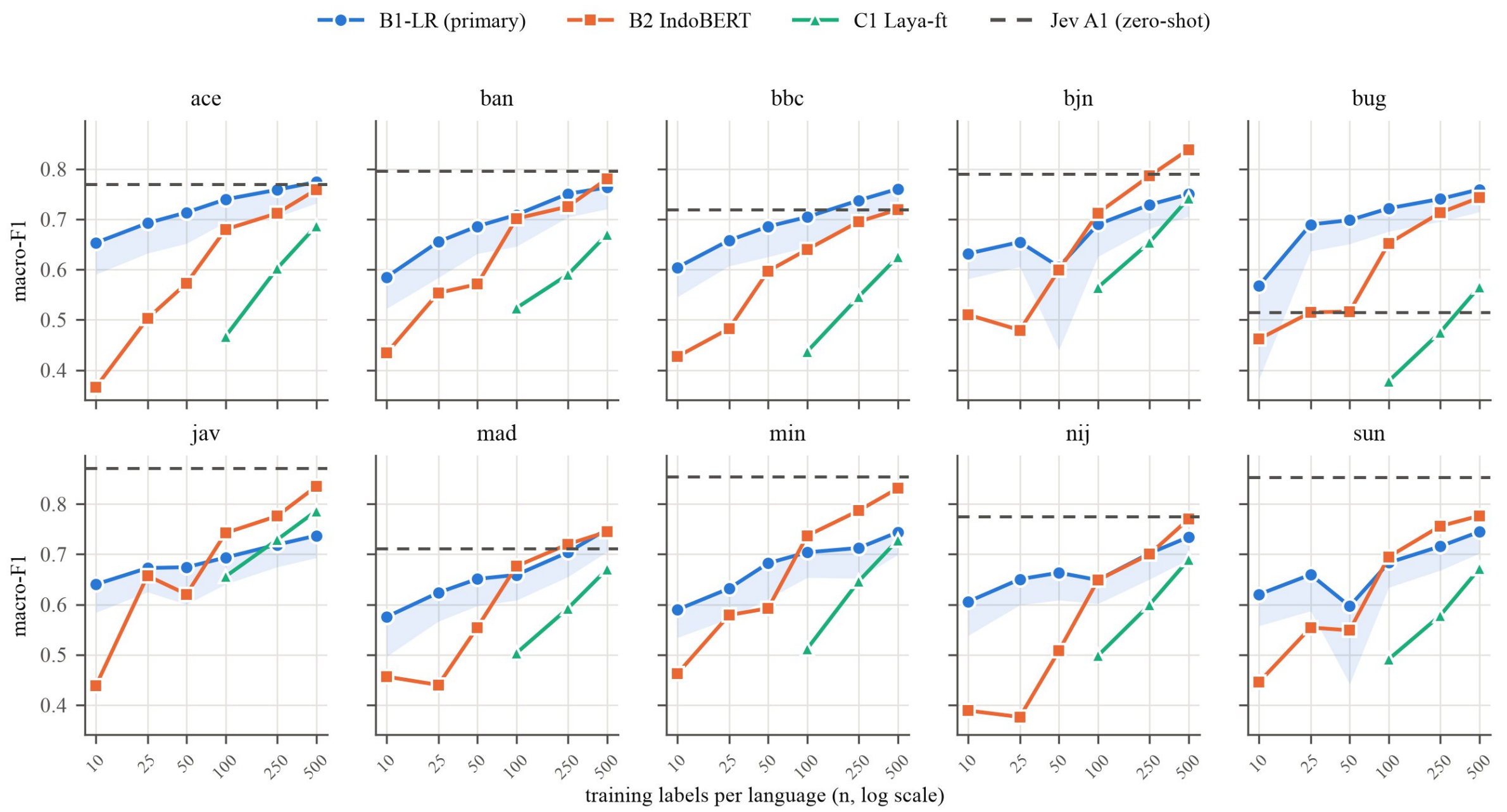

**Fig. 3** Learning curves on NusaX regional languages: macro-F1 of the trained models (mean of three seeds) against zero-shot Jev (dashed). The band shows the classical baseline down to its pooled 95% lower bound. C1 points at $n \leq 50$ are omitted (no effective training; Sect. 3.6)

**Table 6** RQ3: label budget $n*$ at which a trained model reaches zero-shot Jev (NusaX regional languages, macro-F1)

| Language | Jev A1 | B1-LR at n = 500: mean / lower bound | n∗ B1-LR | n∗ B2 | n∗ C1 |
|---|---|---|---|---|---|
| Acehnese | 0.770 | 0.774 / 0.732 | > 500 | > 500 | > 500 |
| Balinese | 0.796 | 0.763 / 0.721 | > 500 | > 500 | > 500 |
| Toba Batak | 0.719 | 0.760 / 0.717 | > 500 | > 500 | > 500 |
| Banjarese | 0.790 | 0.750 / 0.706 | > 500 | 500 | > 500 |
| Buginese | 0.514 | 0.759 / 0.715 | 25 | 100 | > 500 |
| Javanese | 0.871 | 0.737 / 0.693 | > 500 | > 500 | > 500 |
| Madurese | 0.711 | 0.750 / 0.705 | > 500 | > 500 | > 500 |
| Minangkabau | 0.854 | 0.744 / 0.699 | > 500 | > 500 | > 500 |

| Language | Jev A1 | B1-LR at n = 500: mean / lower bound | n∗ B1-LR | n∗ B2 | n∗ C1 |
|---|---|---|---|---|---|
| Ngaju | 0.775 | 0.734 / 0.689 | > 500 | > 500 | > 500 |
| Sundanese | 0.853 | 0.745 / 0.702 | > 500 | > 500 | > 500 |

*Note.* Lower bound: 2.5th percentile of 3 seeds × 2,000 bootstrap draws. Spearman ρ(n∗, Jev F1): B1-LR 0.52 ($p = 0.061$), B2 0.39 ($p = 0.133$); not defined for C1 (no variation).

### 4.5 RQ4: confidence-gated cascades

Table 7 and Fig. 4 show the cascades. At the share r chosen on validation, 8 of 9 cascades are at least as accurate as the better of their two components. Routing the 30% least confident emotion decisions to IndoBERT raises macro-F1 from 0.629 (Jev) and 0.647 (IndoBERT) to 0.676 (+2.9 points over IndoBERT, [1.2, 4.5]), because the two models err on different items. The local fallback adds no API cost and 5.9 ms (IndoBERT) or 79 ms (Laya) per routed decision on a laptop GPU. Routing to DeepSeek instead matches DeepSeek alone at 24–77% lower API cost. The exception is NT-senti → Laya-ft (local): the validation pilot, where Laya was weak, selected too small a share for the test split, where Laya is stronger. The test curves peak at larger shares (30–45%) than those chosen on validation; we do not claim these peaks, as they were selected after seeing the test results.

The cascade does not make a local model cheaper: Jev is called for every item, so a team that has already trained a local classifier pays the Jev cost in exchange for accuracy. Its value lies elsewhere: a model trained once can be kept in reserve for the 5–30% of decisions on which Jev is least sure, and the cascade is at least as accurate as either model alone.

**Table 7** RQ4: cascades at the operating point r chosen on validation (test macro-F1)

| Dataset | Fallback | r | Jev | Fallback | Cascade | Δ vs better [95% CI] | USD / 1,000 | ms |
|---|---|---|---|---|---|---|---|---|
| NusaX-senti | IndoBERT (local) | 10% | 0.782 | 0.792 | **0.801** | +0.9 [−2.0, 3.8] | 0.017 | 404 |
| NusaX-senti | Laya-ft (local) | 5% | 0.782 | 0.786 | **0.795** | +0.9 [−1.8, 3.6] | 0.017 | 407 |
| NusaX-senti | DeepSeek (API) | 60% | 0.782 | 0.786 | **0.789** | +0.2 [0.1, 0.4] | 0.078 | 926 |
| NT-senti | IndoBERT (local) | 15% | 0.781 | 0.824 | **0.820** | −0.4 [−2.0, 1.0] | 0.015 | 379 |
| NT-senti | Laya-ft (local) | 5% | 0.781 | 0.815 | **0.795** | −2.1 [−3.8, −0.4] | 0.015 | 382 |
| NT-senti | DeepSeek (API) | 5% | 0.781 | 0.795 | **0.787** | −0.8 [−1.8, 0.2] | 0.019 | 418 |
| NT-emot | IndoBERT (local) | 30% | 0.629 | 0.647 | **0.676** | +2.9 [1.2, 4.5] | 0.019 | 410 |
| NT-emot | Laya-ft (local) | 5% | 0.629 | 0.611 | **0.636** | +0.7 [0.4, 1.1] | 0.019 | 412 |
| NT-emot | DeepSeek (API) | 30% | 0.629 | 0.658 | **0.662** | +0.4 [−0.7, 1.5] | 0.055 | 675 |

*Note.* Δ: cascade minus the better single model of the pair, in macro-F1 points. Latency: mean per decision (Jev plus the fallback for routed items); local fallbacks measured on a laptop RTX 3050 Ti.

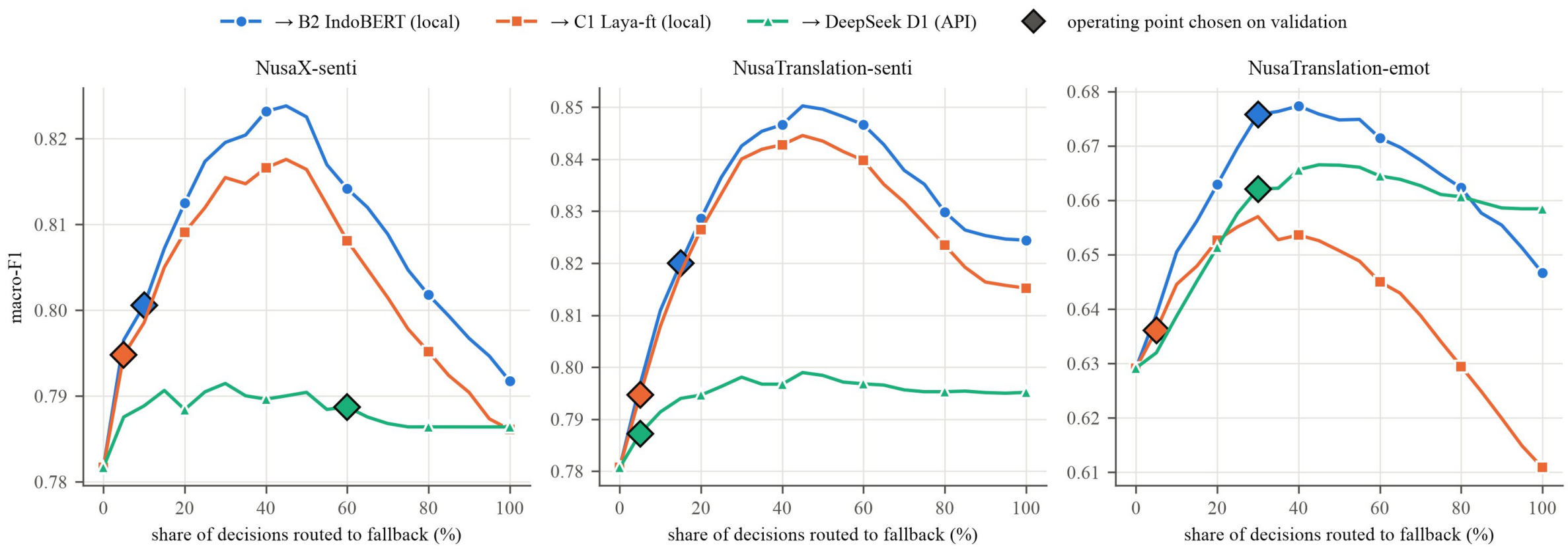

**Fig. 4** Macro-F1 of the cascade as a function of the share of Jev decisions routed to the fallback; diamonds mark the operating point chosen on the validation pilot

### 4.6 Commercial LLMs: DeepSeek, Claude and GPT

Two kinds of LLM comparison were pre-registered. DeepSeek (D1) is the primary LLM comparator: it runs on the full test sets and returns token probabilities, so its accuracy and calibration can be compared with Jev item by item. Claude and GPT form a secondary ladder: they were queried through their command-line clients on the fixed test subsets, and their confidences are verbalised. On the full test sets, Jev is statistically indistinguishable from DeepSeek in 24 of 26 language versions. Table 8 and Fig. 5 report the whole ladder on identical subset items.

**Table 8** LLM ladder on the fixed test subsets: macro-F1, accuracy difference to Jev, API cost and median latency

| Model | NusaX-senti | NT-senti | NT-emot | Δ accuracy vs Jev (points) | USD / 1,000 | Latency p50 (s) |
|---|---|---|---|---|---|---|
| Jev (*noul_en*) | 0.783 | 0.783 | 0.640 | — / — / — | 0.015–0.019 | 0.4 |

| Model | NusaX-senti | NT-senti | NT-emot | Δ accuracy vs Jev (points) | USD / 1,000 | Latency p50 (s) |
|---|---|---|---|---|---|---|
| Jev (*choice_en*) | 0.790 | 0.790 | 0.639 | +0.4 / +0.3 / +0.1 | 0.016–0.018 | 0.4 |
| DeepSeek V4.1 Flash | 0.784 | 0.821 | 0.660 | −0.7 / **+3.3** / +1.7 | 0.085–0.119 | 0.8–0.9 |
| Claude Haiku 4.5 | 0.729 | 0.750 | 0.627 | **−6.5** / **−4.4** / −1.6 | 6.3–7.8 | 8.8–11.3 |
| Claude Sonnet 5 | 0.862 | 0.859 | 0.724 | **+7.8** / **+6.5** / **+9.1** | 5.9–9.4 | 3.1–5.0 |
| Claude Opus 5.5 | 0.890 | 0.877 | 0.770 | **+10.2** / **+7.5** / **+13.6** | 9.4–10.9 | 4.2–4.7 |
| GPT-6 Luna | 0.832 | 0.830 | 0.714 | **+4.2** / **+4.6** / **+8.0** | not reported | 4.1–5.4 |
| GPT-6 Sol | 0.867 | 0.866 | 0.726 | **+8.1** / **+6.8** / **+9.4** | not reported | 4.6–7.0 |
| GPT-5.6 Terra | 0.843 | 0.842 | 0.702 | **+6.0** / **+4.6** / **+6.7** | not reported | 4.4–5.3 |
| IndoBERT (B2), trained | 0.807 | 0.824 | 0.659 | — / — / — | local | — |
| XLM-R (B3), trained | 0.797 | 0.828 | 0.666 | — / — / — | local | — |
| Laya-ft (C1), trained | 0.799 | 0.824 | 0.632 | — / — / — | local | — |

*Note.* Δ: model minus Jev (*noul_en*) in accuracy points for NusaX-senti / NT-senti / NT-emot; bold: Holm-adjusted $p < 0.05$ (exact McNemar, 24 tests). Claude costs are the API-equivalent costs reported by Claude Code; Codex reports no cost. LLM latencies include the command-line client and are not API latencies. Trained models (three seeds) are shown for reference.

The ladder separates into three groups. Claude Opus 5.5 is the most accurate model on every subset (macro-F1 0.890, 0.877, 0.770 against Jev's 0.783, 0.783, 0.640), followed by GPT-6 Sol (0.867, 0.866, 0.726) and Claude Sonnet 5; GPT-6 Luna and GPT-5.6 Terra come next. All five are significantly more accurate than Jev on every subset, by 4.2–13.6 accuracy points, with the largest gaps on the five-class emotion task, and they also exceed the encoders fine-tuned on all available labels. Claude Haiku 4.5, the smallest model, is less accurate than Jev on both sentiment subsets and not significantly different on emotion. DeepSeek, the cheapest LLM, is close to Jev; its one significant subset difference is on NT-senti.

The verbalised confidences of the frontier models rank their errors well: their AURC on NusaX is 0.021–0.033 against 0.066 for Jev's model probabilities. This is consistent with verbalised confidence being informative for strong LLMs [27], but the two kinds of confidence are not compared statistically, as pre-registered: part of the difference reflects the higher accuracy, which leaves fewer errors to rank.

The accuracy has a price. Where cost is reported, Claude Sonnet and Opus cost 347–618 times as much as Jev per decision, and each call took seconds rather than about 369 ms (partly client overhead). The GPT models were used through a subscription that reports no cost, so their price is not compared. Jev therefore sits at the cheap end of the cost–accuracy frontier, and DeepSeek one step above it. Where a few points of accuracy justify a hundredfold cost, a frontier LLM is the better single model; otherwise Jev, alone or with a local fallback (Table 7), is the economical choice. A cascade from Jev to a frontier LLM follows the same logic as the DeepSeek cascade but was not part of our design.

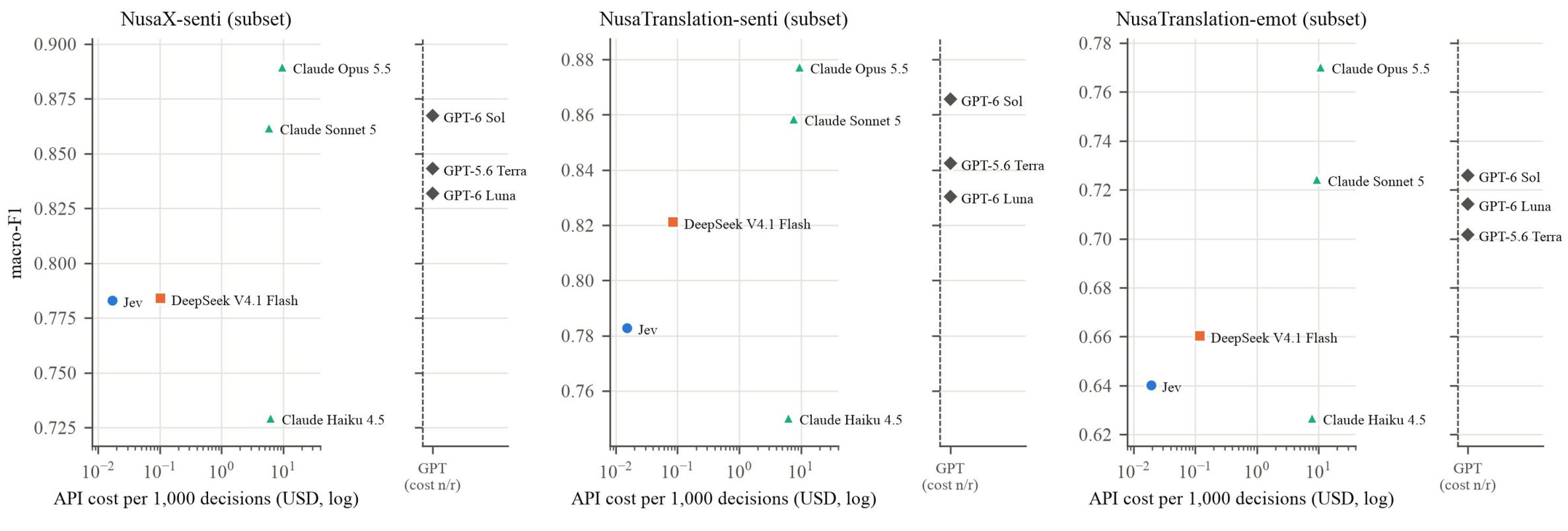

**Fig. 5** Macro-F1 against API cost per 1,000 decisions on the fixed test subsets (log scale). Claude costs are the API-equivalent costs reported by the client; the GPT models report no cost and are shown in the strip to the right of each panel, on the same vertical scale

### 4.7 Translate-test and question formulation

Machine translation is not a substitute for the human Indonesian sentence. Translating with NLLB before asking Jev (A3) lowers accuracy in 6 of 13 language versions (uncorrected $p < 0.05$; descriptive analysis). The exception is Buginese, the language with the lowest NLLB quality (chrF 34.7) and the weakest Jev, where translation adds 11.7 points. Where Jev already understands the regional language, translation mostly adds noise; where it does not, even imperfect translation helps. The question formulation matters little: the three pre-registered formulations differ by at most 1.1 macro-F1 points and agree on most items (Cohen's κ 0.87–0.95), consistent with the option-name findings of Sun et al. [13] once option names are held fixed.

### 4.8 Exploratory analyses

The following analyses were not pre-registered. Across the 24 regional language versions, Jev's macro-F1 correlates with NLLB translation quality ($\rho = 0.61$, $n = 13$) and with FLORES-200 coverage ($\rho = 0.47$), and negatively with genetic distance from Indonesian ($\rho = -0.38$) [41]. Jev's advantage over IndoBERT is largest in languages present in the CC100 pretraining corpus ($\rho = 0.64$), i.e. Javanese and Sundanese. On NT-emot, Jev's recall is lowest for sadness (0.51) and fear (0.51); 25% of sad sentences are labelled as anger, almost exactly as by DeepSeek (25%), while IndoBERT recognises fear better (0.66). Confusions among negative emotions are shared by all systems and may partly reflect the labels.

### 4.9 Implications for practice

Three practical conclusions follow for sentiment and emotion decisions in Indonesian regional languages. First, zero-shot Jev is a credible default where no labelled data exist, at 1.5–1.9 US cents per 1,000 decisions; it matches models fine-tuned on hundreds of labels per language on NusaX. Second, its reliability depends on the language: for languages far from Indonesian and absent from pretraining corpora (Makassarese, Batak, Madurese and Buginese), a few dozen to a few hundred labels or the Indonesian original change the picture, and such languages should be validated before deployment. Third, its probabilities can be used as calibrated scores, but a threshold tuned on Indonesian will remove fewer errors on regional text; a cascade to a local model, with the share chosen on held-out data, is a safer way to use the confidence.

### 4.10 Threats to validity

*Construct.* Sentiment and emotion are two task families, and the regional sentences are translations of Indonesian originals, which may be closer to Indonesian than naturally written regional text. The results should not be generalised to other tasks or to spontaneous regional writing. *Contamination.* NusaX has been public since 2022, and the closed models may have seen it [42]; Jev cannot be probed for memorisation because it does not generate text. NusaTranslation is more recent, and the conclusions agree across the datasets. *Model drift.* Jev is a service; all answers came from one dated snapshot, and a later version may behave differently. The open Laya and the trained baselines make the local-model results reproducible. *Statistics.* All comparisons are paired on identical items, with Holm correction within families; the H3a test has low power (Sect. 4.4), and exploratory correlations rest on at most 24 language versions. *Setting.* The LLM ladder ran through command-line clients on subsets, with verbalised confidences that are not comparable with model probabilities, and local latencies were measured on a single laptop GPU.

## 5 Conclusion

We evaluated calibrated typed-decision models on Indonesian regional languages with a pre-registered, parallel design. Zero-shot Jev is usable: on NusaX sentiment it matches encoders fine-tuned on all available labels, and classical models need more than 500 labels per language to match it in 9 of 10 languages. Its accuracy, however, follows the closeness of each sentence to Indonesian, and the Indonesian original recovers much of what is lost in distant languages, whereas machine translation mostly does not. Its probabilities stay calibrated on regional languages, contrary to our prediction, but they rank its own errors less well, so confidence thresholds transfer less than calibration suggests. A cascade that returns Jev's least confident decisions to a local classifier is at least as accurate as either model and, for emotion classification, better than both (+2.9 macro-F1 points). Frontier Claude and GPT models are clearly more accurate than Jev, but at several hundred times its cost where cost is reported; Jev is the economical first layer rather than a replacement for them.

The study is bounded by its tasks (sentiment and emotion), by translated rather than spontaneous regional text, by one dated snapshot of a commercial service, and by the low power of the label-budget test (Sect. 4.10). Future work should extend the evaluation to other decision types and to naturally written regional text, test whether confidence thresholds can be recalibrated per language with a few labels, and measure cascades in a deployed setting with real traffic and latency.

## Statements and Declarations

**Funding.** This research received no external funding; all costs were covered personally by the authors.

**Competing Interests.** The authors have no competing interests to declare that are relevant to the content of this article.

**Ethics Approval.** Not applicable. The study uses only publicly released, anonymised text datasets and involves no human participants.

**Consent.** Not applicable.

**Availability of Data and Materials.** The NusaX and NusaWrites datasets are publicly available under CC BY-SA 4.0 [9], [10]. The pre-registration, the fixed subsets and learning-curve splits, all model predictions and all result tables are available at https://github.com/tnromy/evaluating-system-one-models-for-indonesian-regional-languages. Jev outputs are released for benchmarking only; the provider's terms prohibit their use as training data [37].

**Code Availability.** All code for data preparation, querying, fine-tuning, analysis and figures, including the tagged pre-registration (prereg-v1), is available at https://github.com/tnromy/evaluating-system-one-models-for-indonesian-regional-languages.

**Authors' Contributions.** R.F.: conceptualisation, methodology, software, investigation, formal analysis, writing (original draft). R.: supervision, methodology, validation, writing (review and editing). W.R.P.U.: validation, data curation, writing (review and editing). Z.M.A.: validation, visualisation, writing (review and editing). All authors read and approved the final manuscript.

**Use of AI Tools.** Claude (Anthropic), used through Claude Code, assisted with programming, statistical analysis scripts and language editing of this manuscript. All study design decisions, analyses and interpretations were checked by the authors, who take full responsibility for the content. The LLMs evaluated as experimental arms (DeepSeek, Claude and GPT models) are described in the Methods and were not used to write the paper.

## References

[1] G. Booch, F. Fabiano, L. Horesh, K. Kate, J. Lenchner, N. Linck, A. Loreggia, K. Murgesan, N. Mattei, F. Rossi, B. Srivastava, "Thinking Fast and Slow in AI," *Proceedings of the AAAI Conference on Artificial Intelligence*, vol. 35, no. 17, pp. 15042–15046, 2021. doi: 10.1609/aaai.v35i17.17765.

[2] D. Almeida, "Introducing System One Models & Jev," TypeSafe AI blog, Sep. 15, 2026. [Online]. Available: https://typesafe.ai/blog/introducing-system-one-models-and-jev (accessed Sep. 23, 2026).

[3] TypeSafe AI, "Models," TypeSafe Documentation, 2026. [Online]. Available: https://docs.typesafe.ai/models (accessed Sep. 23, 2026).

[4] Convai Innovations, "Laya Multilingual," Hugging Face model card, revision e4e9ddf, 2026. [Online]. Available: https://huggingface.co/convaiinnovations/laya-multilingual (accessed Sep. 23, 2026).

[5] H. Ibrahim, Y. Zaki, "Evaluating Decision Models for Text Annotation in Computational Social Science," 2026. arXiv:2609.24574. [Online]. Available: https://arxiv.org/abs/2609.24574.

[6] T. Deußer, L. Sparrenberg, R. Sifa, "Evaluating and Benchmarking the System One Model Jev," 2026. arXiv:2609.37647. [Online]. Available: https://arxiv.org/abs/2609.37647.

[7] A. F. Aji, G. I. Winata, F. Koto, S. Cahyawijaya, A. Romadhony, R. Mahendra, K. Kurniawan, D. Moeljadi, R. E. Prasojo, T. Baldwin, J. H. Lau, S. Ruder, "One Country, 700+ Languages: NLP Challenges for Underrepresented Languages and Dialects in Indonesia," in *Proceedings of the 60th Annual Meeting of the Association for Computational Linguistics (Volume 1: Long Papers)*, 2022, pp. 7226–7249. doi: 10.18653/v1/2022.acl-long.500.

[8] P. Joshi, S. Santy, A. Budhiraja, K. Bali, M. Choudhury, "The State and Fate of Linguistic Diversity and Inclusion in the NLP World," in *Proceedings of the 58th Annual Meeting of the Association for Computational Linguistics*, 2020, pp. 6282–6293. doi: 10.18653/v1/2020.acl-main.560.

[9] G. I. Winata, A. F. Aji, S. Cahyawijaya, R. Mahendra, F. Koto, A. Romadhony, K. Kurniawan, D. Moeljadi, R. E. Prasojo, P. Fung, T. Baldwin, J. H. Lau, R. Sennrich, S. Ruder, "NusaX: Multilingual Parallel Sentiment Dataset for 10 Indonesian Local Languages," in *Proceedings of the 17th Conference of the European Chapter of the Association for Computational Linguistics*, 2023, pp. 815–834. doi: 10.18653/v1/2023.eacl-main.57.

[10] S. Cahyawijaya, H. Lovenia, F. Koto, D. Adhista, E. Dave, S. Oktavianti, S. Akbar, J. Lee, N. Shadieq, T. W. Cenggoro, H. Linuwih, B. Wilie, G. Muridan, G. Winata, D. Moeljadi, A. F. Aji, A. Purwarianti, P. Fung, "NusaWrites: Constructing High-Quality Corpora for Underrepresented and Extremely Low-Resource Languages," in *Proceedings of the 13th International Joint Conference on Natural Language Processing and the 3rd Conference of the Asia-Pacific Chapter of the Association for Computational Linguistics (Volume 1: Long Papers)*, 2023, pp. 921–945. doi: 10.18653/v1/2023.ijcnlp-main.60.

[11] AHTOOOXA, "jev-cyrillic-audit: Paired Russian–English Audit of Jev Accuracy and Calibration," 2026. GitHub repository. [Online]. Available: https://github.com/AHTOOOXA/jev-cyrillic-audit (accessed Sep. 24, 2026).

[12] M. Marone, O. Weller, W. Fleshman, E. Yang, D. J. Lawrie, B. V. Durme, "mmBERT: A Modern Multilingual Encoder with Annealed Language Learning," 2025. arXiv:2509.06888. [Online]. Available: https://arxiv.org/abs/2509.06888.

[13] Y. Sun, J. Xu, J. Shi, Z. Yang, "Type-Safe Is Not Error-Free: A Constrained Decision Head Follows the Option Name, Not the Rubric Bound to It," 2026. arXiv:2609.26758. [Online]. Available: https://arxiv.org/abs/2609.26758.

[14] B. Wilie, K. Vincentio, G. I. Winata, S. Cahyawijaya, X. Li, Z. Y. Lim, S. Soleman, R. Mahendra, P. Fung, S. Bahar, A. Purwarianti, "IndoNLU: Benchmark and Resources for Evaluating Indonesian Natural Language Understanding," in *Proceedings of the 1st Conference of the Asia-Pacific Chapter of the Association for Computational Linguistics and the 10th International Joint Conference on Natural Language Processing*, 2020, pp. 843–857. doi: 10.18653/v1/2020.aacl-main.85.

[15] F. Koto, A. Rahimi, J. H. Lau, T. Baldwin, "IndoLEM and IndoBERT: A Benchmark Dataset and Pre-trained Language Model for Indonesian NLP," in *Proceedings of the 28th International Conference on Computational Linguistics*, 2020, pp. 757–770. doi: 10.18653/v1/2020.coling-main.66.

[16] V. D. Lai, N. Ngo, A. Pouran Ben Veyseh, H. Man, F. Dernoncourt, T. Bui, T. H. Nguyen, "ChatGPT Beyond English: Towards a Comprehensive Evaluation of Large Language Models in Multilingual Learning," in *Findings of the Association for Computational Linguistics: EMNLP 2023*, 2023, pp. 13171–13189. doi: 10.18653/v1/2023.findings-emnlp.878.

[17] A. Lauscher, V. Ravishankar, I. Vulic, G. Glavas, "From Zero to Hero: On the Limitations of Zero-Shot Language Transfer with Multilingual Transformers," in *Proceedings of the 2020 Conference on Empirical Methods in Natural Language Processing (EMNLP)*, 2020, pp. 4483–4499. doi: 10.18653/v1/2020.emnlp-main.363.

[18] W. Zhang, Y. Deng, B. Liu, S. Pan, L. Bing, "Sentiment Analysis in the Era of Large Language Models: A Reality Check," in *Findings of the Association for Computational Linguistics: NAACL 2024*, 2024, pp. 3881–3906. doi: 10.18653/v1/2024.findings-naacl.246.

[19] M. Bucher, M. Martini, "Fine-Tuned 'Small' LLMs (Still) Significantly Outperform Zero-Shot Generative AI Models in Text Classification," 2024. arXiv:2406.08660. [Online]. Available: https://arxiv.org/abs/2406.08660.

[20] T. Le Scao, A. Rush, "How many data points is a prompt worth?," in *Proceedings of the 2021 Conference of the North American Chapter of the Association for Computational Linguistics: Human Language Technologies*, 2021, pp. 2627–2636. doi: 10.18653/v1/2021.naacl-main.208.

[21] T. Ermakova, B. Fabian, E. Golimblevskaia, M. Henke, "A Comparison of Commercial Sentiment Analysis Services," *SN Computer Science*, vol. 4, no. 5, 2023. doi: 10.1007/s42979-023-01886-y.

[22] M. Pakdaman Naeini, G. Cooper, M. Hauskrecht, "Obtaining Well Calibrated Probabilities Using Bayesian Binning," *Proceedings of the AAAI Conference on Artificial Intelligence*, vol. 29, no. 1, 2015. doi: 10.1609/aaai.v29i1.9602.

[23] C. Guo, G. Pleiss, Y. Sun, K. Q. Weinberger, "On Calibration of Modern Neural Networks," in *Proceedings of the 34th International Conference on Machine Learning (ICML)*, 2017.

[24] T. Gneiting, A. E. Raftery, "Strictly Proper Scoring Rules, Prediction, and Estimation," *Journal of the American Statistical Association*, vol. 102, no. 477, pp. 359–378, 2007. doi: 10.1198/016214506000001437.

[25] K. Ahuja, S. Sitaram, S. Dandapat, M. Choudhury, "On the Calibration of Massively Multilingual Language Models," in *Proceedings of the 2022 Conference on Empirical Methods in Natural Language Processing*, 2022, pp. 4310–4323. doi: 10.18653/v1/2022.emnlp-main.290.

[26] S. Kadavath, T. Conerly, A. Askell, T. Henighan, D. Drain, E. Perez, N. Schiefer, Z. Dodds, N. Dassarma, E. Tran-Johnson, S. Johnston, S. El-Showk, A. Jones, N. Elhage, T. Hume, A. Chen, Y. Bai, S. Bowman, S. Fort, D. Ganguli, D. Hernandez, J. Jacobson, J. Kernion, S. Kravec, L. Lovitt, K. Ndousse, C. Olsson, S. Ringer, D. Amodei, T. B. Brown, J. Clark, N. Joseph, B. Mann, S. McCandlish, C. Olah, J. Kaplan, "Language Models (Mostly) Know What They Know," 2022. arXiv:2207.05221. [Online]. Available: https://arxiv.org/abs/2207.05221.

[27] M. Xiong, Z. Hu, X. Lu, Y. Li, J. Fu, J. He, B. Hooi, "Can LLMs Express Their Uncertainty? An Empirical Evaluation of Confidence Elicitation in LLMs," in *International Conference on Learning Representations (ICLR)*, 2024.

[28] Y. Geifman, G. Uziel, R. El-Yaniv, "Bias-Reduced Uncertainty Estimation for Deep Neural Classifiers," in *International Conference on Learning Representations (ICLR)*, 2019.

[29] M. Artetxe, V. Goswami, S. Bhosale, A. Fan, L. Zettlemoyer, "Revisiting Machine Translation for Cross-lingual Classification," in *Proceedings of the 2023 Conference on Empirical Methods in Natural Language Processing*, 2023, pp. 6489–6499. doi: 10.18653/v1/2023.emnlp-main.399.

[30] NLLB Team, M. R. Costa-jussà, J. Cross, O. Çelebi, M. Elbayad, K. Heafield, K. Heffernan, E. Kalbassi, J. Lam, D. Licht, J. Maillard, A. Sun, S. Wang, G. Wenzek, A. Youngblood, B. Akula, L. Barrault, G. M. Gonzalez, P. Hansanti, J. Hoffman, S. Jarrett, K. R. Sadagopan, D. Rowe, S. Spruit, C. Tran, P. Andrews, N. F. Ayan, S. Bhosale, S. Edunov, A. Fan, C. Gao, V. Goswami, F. Guzmán, P. Koehn, A. Mourachko, C. Ropers, S. Saleem, H. Schwenk, J. Wang, "Scaling neural machine translation to 200 languages," *Nature*, vol. 630, no. 8018, pp. 841–846, 2024. doi: 10.1038/s41586-024-07335-x.

[31] L. Chen, M. Zaharia, J. Y. Zou, "FrugalGPT: How to Use Large Language Models While Reducing Cost and Improving Performance," *Transactions on Machine Learning Research*, 2024.

[32] Y. Li, Y. Miao, R. Krishnan, R. Padman, "JEV-as-a-Judge: Accept When Confident, Escalate When Unsure," 2026. arXiv:2609.26550. [Online]. Available: https://arxiv.org/abs/2609.26550.

[33] DeepSeek-AI, "DeepSeek-V4: Towards Highly Efficient Million-Token Context Intelligence," 2026. arXiv:2606.19348. [Online]. Available: https://arxiv.org/abs/2606.19348.

[34] Anthropic, "System Card: Claude Sonnet 5," Jun. 30, 2026. [Online]. Available: https://www.anthropic.com/claude-sonnet-5-system-card (accessed Sep. 23, 2026).

[35] F. Pedregosa, G. Varoquaux, A. Gramfort, V. Michel, B. Thirion, O. Grisel, M. Blondel, G. Louppe, P. Prettenhofer, R. Weiss, R. J. Weiss, J. VanderPlas, A. Passos, D. Cournapeau, M. Brucher, M. Perrot, E. Duchesnay, "Scikit-learn: Machine Learning in Python," *Journal of Machine Learning Research*, vol. 12, pp. 2825–2830, 2011.

[36] A. Conneau, K. Khandelwal, N. Goyal, V. Chaudhary, G. Wenzek, F. Guzman, E. Grave, M. Ott, L. Zettlemoyer, V. Stoyanov, "Unsupervised Cross-lingual Representation Learning at Scale," in *Proceedings of the 58th Annual Meeting of the Association for Computational Linguistics*, 2020, pp. 8440–8451. doi: 10.18653/v1/2020.acl-main.747.

[37] TypeSafe AI, "Master Customer Agreement," 2026. [Online]. Available: https://typesafe.ai/legal/mca (accessed Sep. 23, 2026).

[38] D. Bates, M. Mächler, B. Bolker, S. Walker, "Fitting Linear Mixed-Effects Models Using lme4," *Journal of Statistical Software*, vol. 67, no. 1, 2015. doi: 10.18637/jss.v067.i01.

[39] T. G. Dietterich, "Approximate Statistical Tests for Comparing Supervised Classification Learning Algorithms," *Neural Computation*, vol. 10, no. 7, pp. 1895–1923, 1998. doi: 10.1162/089976698300017197.

[40] E. Miller, "Adding Error Bars to Evals: A Statistical Approach to Language Model Evaluations," 2024. arXiv:2411.00640. [Online]. Available: https://arxiv.org/abs/2411.00640.

[41] P. Littell, D. R. Mortensen, K. Lin, K. Kairis, C. Turner, L. Levin, "URIEL and lang2vec: Representing languages as typological, geographical, and phylogenetic vectors," in *Proceedings of the 15th Conference of the European Chapter of the Association for Computational Linguistics: Volume 2, Short Papers*, 2017, pp. 8–14.

[42] S. Balloccu, P. Schmidtova, M. Lango, O. Dusek, "Leak, Cheat, Repeat: Data Contamination and Evaluation Malpractices in Closed-Source LLMs," in *Proceedings of the 18th Conference of the European Chapter of the Association for Computational Linguistics (Volume 1: Long Papers)*, 2024, pp. 67–93. doi: 10.18653/v1/2024.eacl-long.5.